\documentclass{ifacconf}

\usepackage{booktabs}

\usepackage{graphicx}      % include this line if your document contains figures
\usepackage{natbib}        % required for bibliography
\usepackage{amsmath}
\usepackage{amsfonts}
\usepackage{mathtools}

\usepackage{thmtools} % for theorem number: main/appendix
\theoremstyle{definition}
\newtheorem{definition}{Definition}[section]

\newcommand{\rBr}[1]{\left(#1\right)}
\newcommand{\sqBr}[1]{\left[#1\right]}

\newcommand{\vecOp}[1]{\operatorname{vec}\left(#1\right)}

\newcommand{\tens}[1]{\boldsymbol{\mathcal{#1}}} % Tensor calligraphic
\newcommand{\mat}[1]{\boldsymbol{#1}} % Matrix bold
\newcommand{\vect}[1]{\boldsymbol{#1}} % Vectors bold
\newcommand{\inCo}[1]{\in\mathbb{C}^{#1}} % \in\Real{}
\begin{document}
\begin{frontmatter}

\title{Laplace Approximation For Tensor Train Kernel Machines In System Identification\thanksref{footnoteinfo}} 

\thanks[footnoteinfo]{This work is supported by the
Dutch Research Council (NWO).}

\author[First]{Albert Saiapin} 
\author[First]{Kim Batselier} 
%\author[Third]{Third C. Author}

\address[First]{Delft Center for Systems and Control, TU Delft, Netherlands (e-mail: \{a.saiapin, k.batselier\}@tudelft.nl)}
%\address[Second]{Colorado State University, 
%   Fort Collins, CO 80523 USA (e-mail: author@lamar. colostate.edu)}
%\address[Third]{Electrical Engineering Department, 
%   Seoul National University, Seoul, Korea, (e-mail: author@snu.ac.kr)}

\begin{abstract} % Abstract of 50--100 words
To address the scalability limitations of Gaussian process (GP) regression, several approximation techniques have been proposed. One such method is based on tensor networks, which utilizes an exponential number of basis functions without incurring exponential computational cost. However, extending this model to a fully probabilistic formulation introduces several design challenges. In particular, for tensor train (TT) models, it is unclear which TT-core should be treated in a Bayesian manner.
We introduce a Bayesian tensor train kernel machine that applies Laplace approximation to estimate the posterior distribution over a selected TT-core and employs variational inference (VI) for precision hyperparameters. 
Experiments show that core selection is largely independent of TT-ranks and feature structure, and that VI replaces cross-validation while offering up to 65× faster training. The method’s effectiveness is demonstrated on an inverse dynamics problem.
\end{abstract}


\end{frontmatter}
%===============================================================================

\section{Introduction}
Nonlinear system identification often relies on black-box models that learn input–output relationships directly from data. Classical approaches typically assume a parametric structure and select model complexity using criteria such as Akaike information criterion, see e.g. \cite{AIC_2019}. While effective for linear systems, these methods often struggle to capture more complex nonlinear behavior. An alternative is to use regression techniques — such as neural models, including radial basis function networks introduced by~\cite{RBF_1988}.%and partition-based approaches like LoLiMoT, introduced by~\cite{BFN_1996}. 

Another popular method of choice for system identification is Gaussian process (GP) regression, see e.g.~\cite{GPR_2003}. This powerful probabilistic framework places a Gaussian process prior over the unknown function and uses observed data to infer a posterior distribution, providing both predictions and principled uncertainty estimates. Its ability to learn complex structures directly from data has led to successful applications in model predictive control e.g. \cite{MPC_GPR_2017}, and nonlinear state estimation - see e.g. \cite{GPR_SI_2019}; \cite{GPR_NSI_TV_2018}. However, despite their flexibility and accuracy, standard GPs scale cubically with the number of data points, which poses a major limitation for large-scale or real-time identification tasks.

\begin{figure}
\begin{center}
\includegraphics[width=8.4cm]{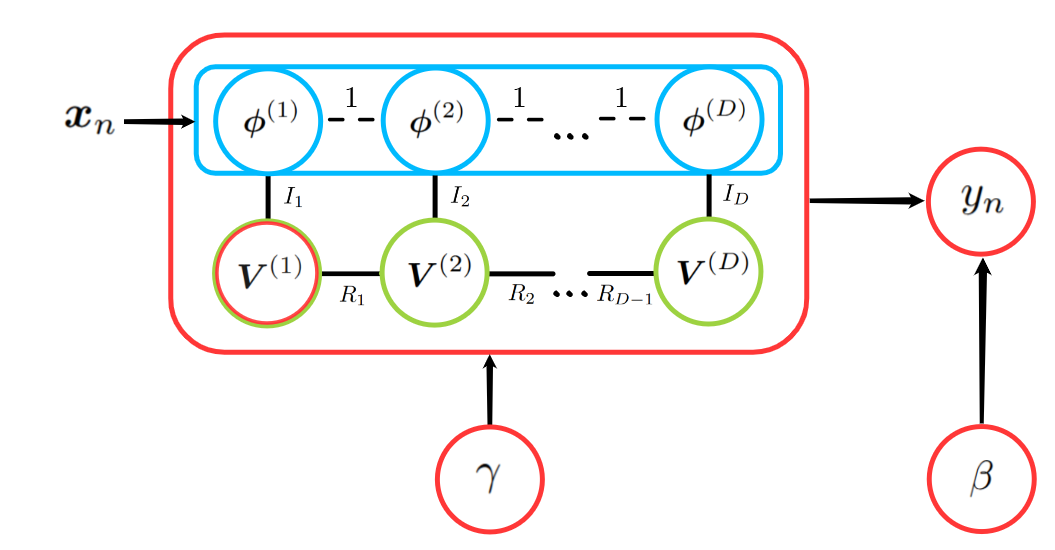}    % The printed column width is 8.4 cm.
\caption{Probabilistic graphical model representing the Bayesian tensor train kernel machine~\eqref{la_ttkm}. Each circle represents a tensor, with its order indicated by the number of outgoing solid lines. Blue color represents 
nonlinear features $\vect{\phi}^{(d)}(x_d)$; green circles correspond to TT model parameters; and red circles denote random variables. Solid lines indicate summation over the corresponding index, while dashed lines denote a Kronecker product~\citep{TN1_Cichocki_2016}.} 
\label{fig:la_ttkm}
\end{center}
\end{figure}

Scalability issues of GPs have led to extensive research on reducing the computational limitations through kernel approximations; see e.g. \cite{GPR_BigData_2018}. 
A particularly influential strategy is to approximate the kernel with a finite set of basis functions, thereby converting the original nonparametric GP into a parametric model, see e.g. \cite{HSM_RR_GPR_2020}. The resulting model enables predictions with complexity $\mathcal{O}(NL^2)$, where $N$ is the number of training points and $L$ is the number of basis functions controlling approximation accuracy. However, if achieving sufficient accuracy requires a large $L$, the computational cost quickly becomes prohibitive.
To overcome this limitation, \cite{GPR_TT_2023} proposed an approach that exploits low-rank tensor networks (TNs) to represent an exponentially large number of basis functions without incurring exponential complexity. Their method first identifies a suitable low-dimensional subspace described by a low-rank TN, performs Bayesian inference to estimate the model weights in that subspace, and finally projects the inferred weights back to the original space for GP predictions.

Our work is inspired by \cite{GPR_TT_2023} and extends it in several important directions. First, we introduce the Bayesian tensor train kernel machine (LA-TTKM), which unifies variational inference with a Laplace approximation applied to a selected TT-core. The resulting probabilistic graphical model is shown in Figure~\ref{fig:la_ttkm}. The key idea is to approximate the local posterior of the model parameters with a Gaussian centered at the mode identified by the optimization procedure, where the covariance captures the local curvature of the likelihood. Although our derivation starts from a different perspective, it leads to a formulation closely related to the original approach.
Second, we resolve the open questions posed in \cite{GPR_TT_2023}. We systematically investigate how the choice of TT-ranks influences model performance across datasets and study which TT-core should be treated in a Bayesian manner. Moreover, we show that variational inference provides a highly efficient alternative to cross-validation for hyperparameter optimization in probabilistic TN-based models, achieving speedups of up to 65× while preserving predictive quality. Finally, we demonstrate the effectiveness and practical relevance of the proposed method on an inverse dynamics estimation task. 

\section{Background}\label{sec:background}
We denote sets with calligraphic capital letters - $\mathcal{Z}$.
Scalars are denoted in italics $w, W$, vectors in lowercase bold $\vect{w}$, indexed vectors as $\vect{w}_s$, matrices in capital bold $\mat{W}$ and tensors, also being high-order arrays, in capital italic bold font $\tens{W}$. The $i$th entry of a vector $\vect{w} \inCo{I}$ is denoted as $w_i$ and the $i_1i_2\dots i_D$th entry of a $D$th-order tensor $\tens{W} \inCo{I_1 \times I_2 \times \dots \times I_D}$ as $w_{i_1 i_2 \dots i_D}$ or $\sqBr{\tens{W}}_{i_1 i_2 \dots i_D}$. 
The conjugate-transpose of $\mat{A}$ is denoted as $\mat{A}^\top$, 
and $\otimes$, $\odot_R$ represent the Kronecker product and row-wise Khatri-Rao product correspondingly~\citep{TN1_Cichocki_2016}. We employ zero-based indexing for all tensors.
Since working with vectors and matrices is more straightforward than with tensors, we introduce the vectorization operation.
\begin{definition}[Vectorization]\label{def_vec}
    The vectorization operator $\vecOp{\cdot}: \mathbb{C}^{I_1 \times I_2 \times \dots \times I_D} \rightarrow \mathbb{C}^{I_1I_2\dots I_D}$ is defined as:
    \begin{equation*}
        \vecOp{\tens{W}}_i \coloneq w_{i_1 i_2 \dots i_D},
    \end{equation*}
    where $i \coloneq \rBr{i_1i_2\dots i_D}= i_1 + \sum_{d=2}^{D}i_d \prod_{j=1}^{d-1}I_j$.
\end{definition}

\subsection{Gaussian process regression}
Gaussian process regression addresses the following task:  
given observed input–output pairs $\mathcal{D} \coloneq \{\rBr{\vect{x}_n, y_n} | \vect{x}_n \in \mathcal{X}^D, y_n \in \mathcal{Y},  n=1, \dots, N\} = \{\mat{X}\in \mathcal{X}^{N\times D}, \vect{y}\in \mathcal{Y}^{N}\}$ generated by an unknown function $y = f(\vect{x})$, the goal is to predict the function values at a new input $\vect{x}_*$.  
The model assumes
\begin{align}
    y_n &= f(\vect{x}_n) + \epsilon_n \text{ , } \epsilon_n \sim \mathcal{N}(0, \beta^{-1}),  \label{eq:gp_func} \\
    f(\vect{x}_n) &\sim \mathcal{GP}(m(\vect{x}), k(\vect{x}, \vect{x}^\prime)), \label{eq:gp_prior}
\end{align} 
where~\eqref{eq:gp_prior} specifies a Gaussian process prior with mean function  
$m(\vect{x}) = \mathbb{E}[f(\vect{x})]$ and covariance (kernel) function 
$k(\vect{x}, \vect{x}') = \operatorname{Cov}(f(\vect{x}), f(\vect{x}'))$.  
Equation~\eqref{eq:gp_func} states that the observations are noisy evaluations of the latent function, with independent Gaussian noise of precision $\beta$.

The posterior distribution of the function value at a new input $\vect{x}_*$ is obtained by conditioning the joint Gaussian prior on the observed outputs $\vect{y}$; see, e.g.,~\cite{GPR_CSD_2022}:
\begin{equation}\label{eq:gp_pred}
    \begin{split}
        f(\vect{x}_*)|\mat{X}, \vect{y} &\sim \mathcal{N}(\mathbb{E}[f(\vect{x}_*)], \mathbb{V}[f(\vect{x}_*)]), \\
        \mathbb{E}[f(\vect{x}_*)] &= \vect{k}_*^\top\sqBr{\mat{K} + \beta^{-1}\mat{I}}^{-1}\vect{y}, \\
        \mathbb{V}[f(\vect{x}_*)] &= k_{**} - \vect{k}_*^\top\sqBr{\mat{K} + \beta^{-1}\mat{I}}^{-1}\vect{k}_*,
    \end{split}
\end{equation}
where the kernel matrices are defined element-wise as $\sqBr{\mat{K}}_{ij} = k(\vect{x}_i, \vect{x}_j)$, $\sqBr{\vect{k}_*}_i = k(\vect{x}_i, \vect{x}_*)$ and $k_{**} = k(\vect{x}_*, \vect{x}_*)$.
The main computational cost arises from inverting the $N \times N$ matrix in~\eqref{eq:gp_pred}, which requires $\mathcal{O}(N^3)$ operations.

\subsection{Tensor network kernel machine}

Consider the following linear regression model:
\begin{equation}\label{intro_simple_model}
    f(\vect{x}) = \vect{\phi}(\vect{x})^\top \vect{w},
\end{equation}
where $\vect{\phi}(\vect{x}) \in \mathbb{C}^{I_1 I_2 \dots I_D}$ denotes a nonlinear feature map and 
$\vect{w} \in \mathbb{C}^{I_1 I_2 \dots I_D}$ is a vector of weights learned from the dataset $\mathcal{D}$. 
%The choice of feature map is crucial, as it enables the model to capture complex nonlinear structure in the data.  
In the setting of tensor network kernel machines, this formulation is accompanied by two additional assumptions:
\begin{enumerate}
    \item Features $\vect{\phi}(\vect{x})$ have tensor-product structure: \begin{equation}\label{tensor_product_features}
    \vect{\phi}(\vect{x}) \coloneq \vect{\phi}^{(D)}(x_D) \otimes \dots \otimes \vect{\phi}^{(1)}(x_1),
    \end{equation}
    where $\vect{\phi}^{(d)}: \mathbb{C} \rightarrow \mathbb{C}^{I_d}$ is a feature map acting on the $d$th component of $\vect{x} \inCo{D}$.
    \item Model weights $\vect{w}$ are represented as a tensor network.
\end{enumerate}
The tensor–product structure in~\eqref{tensor_product_features} is closely connected to product kernels, see e.g.~\cite{HSM_RR_GPR_2020}. At first glance, a major drawback of this construction is that the input $\vect{x}$ is mapped to an exponentially large feature vector 
$\vect{\phi}(\vect{x}) \in \mathbb{C}^{I_1 I_2 \dots I_D}$,  
which in turn implies an exponential number of parameters in $\vect{w}$.  
This observation motivates the second assumption, which introduces a low-rank structure on the model parameters.

%TT_Volterra_2017

In this paper we focus on tensor train (TT) network as a model parameterization, see e.g.~\cite{TT_2011, NSI_TN_2022}.
\begin{definition}[Tensor train (TT)]\label{def_tt}
    A Dth-order tensor $\tens{W} \in \mathbb{C}^{I_1 \times I_2 \times \dots \times I_D}$ is represented as a tensor train with rank-$(R_1, R_2,\dots , R_{D-1})$ if 
    \begin{equation*}
        \mathcal{W}_{i_1\dots i_D} = \mat{V}^{\rBr{1}}\rBr{i_1}\dots\mat{V}^{\rBr{D}}\rBr{i_D},
    \end{equation*}
    where $\tens{V}^{\rBr{d}} \inCo{R_{d-1}\times I\times R_d}$ are called TT-cores, $\mat{V}^{\rBr{d}}\rBr{i_d} \coloneq \sqBr{\tens{V}^{\rBr{d}}}_{:i_d:} \inCo{R_{d-1}\times R_d}$, and $R_0 = R_D \coloneq 1$, $d=1, \dots, D$.
\end{definition}

Training a tensor train kernel machine means solving the following nonlinear nonconvex optimization problem:
\begin{equation}\label{tt_km_opt}
    \begin{split}
        &\min_{\vect{w}}\frac{1}{2} \|\vect{y} - \mat{\Phi}\vect{w}\|_2^2 + \frac{\alpha}{2}\|\vect{w}\|_2^2 \\
        & \text{s.t. }w_{\rBr{i_1i_2\dots i_D}} = \mat{V}^{\rBr{1}}\rBr{i_1}\dots\mat{V}^{\rBr{D}}\rBr{i_D},
    \end{split} 
\end{equation}
where $\vect{y}$ is the target variable vector, $\Phi_{ni} \coloneq \phi_i(\vect{x}_n)$, $\vect{x}_n$ represents an $n$th row of data matrix $\mat{X} \inCo{N\times D}$, and $\alpha$ is a regularization hyperparameter.
Note that, to compute the model response $\vect{f} \coloneq \mat{\Phi} \vect{w}$, we implicitly take its real part, as the features $\vect{\phi}$ and weights $\vect{w}$ are generally complex.

The optimal model parameters $\tens{V}^{\rBr{d}}$ in the minimization problem~\eqref{tt_km_opt} can be found using the ALS algorithm~\citep{TT_Volterra_2017}, where each TT core $\tens{V}^{(d)}$ is updated sequentially for $d=1, \dots, D, \dots, 2$ while keeping the other cores fixed:
\begin{equation}\label{eq:als}
    \begin{split}
        &\vect{v}^{\rBr{d}} \coloneq \vecOp{\tens{V}^{\rBr{d}}} = \rBr{\mat{A}^{(d)\top} \mat{A}^{(d)} + \dfrac{\gamma}{\beta} \mat{I}}^{-1}\mat{A}^{(d)\top}\vect{y},\\
        &\mat{A}^{\rBr{d}}=\begin{cases}
            &\mat{Q}^{\rBr{d}} \odot_R \mat{\Phi}^{\rBr{d}} \text{, if $d=1$},\\
            &\mat{\Phi}^{\rBr{d}} \odot_R \mat{P}^{\rBr{d}} \text{, if $d=D$},\\
            &\mat{Q}^{\rBr{d}} \odot_R \mat{\Phi}^{\rBr{d}} \odot_R \mat{P}^{\rBr{d}} \text{, else}.
        \end{cases}
    \end{split}
\end{equation}
with the auxiliary matrices defined as follows:
\begin{equation}
    \begin{split}
        &\sqBr{\mat{\Phi}^{(d)}}_{ni} \coloneq \vect{\phi}_{i}^{(d)}(x_{nd}), \\
        &\begin{array}{ll}
        \sqBr{\mat{P}^{\rBr{d}}}_{nr_{d-1}} \coloneq &\sum_{r_1\dots r_{d-2}} \sqBr{\sum_{i_1} \Phi_{ni_1}^{\rBr{1}} V_{i_1 r_1}^{\rBr{1}}} \dots \\
        & \dots \sqBr{\sum_{i_{d-1}} \Phi_{ni_{d-1}}^{\rBr{d-1}} V_{r_{d-2} i_{d-1} r_{d-1}}^{\rBr{d-1}}},
        \end{array}\\
        &\begin{array}{ll}
        \sqBr{\mat{Q}^{\rBr{d}}}_{nr_d} \coloneq \sum_{r_{d+1}\dots r_{D-1}} &\sqBr{\sum_{i_{d+1}} \Phi_{ni_{d+1}}^{\rBr{d+1}} V_{r_d i_{d+1} r_{d+1}}^{\rBr{d+1}}} \dots \\
        & \dots \sqBr{\sum_{i_{D}} \Phi_{ni_{D}}^{\rBr{D}} V_{r_{D-1} i_{D}}^{\rBr{D}}}.
        \end{array}
    \end{split}
\end{equation}
The ALS update is done for several epochs (sweeps) or before the target loss goes down sufficiently enough. Training time and memory complexities of the algorithm are  $T  = \mathcal{O}\rBr{EDI^2R^4\rBr{N+IR^2}}$ and $M = \mathcal{O}\rBr{NR\rBr{D + IR}}$ respectively, where $E$ represents the number of ALS epochs, $R = \max_{d=1, \dots, D-1}R_d$, and $I=\max_{d=1, \dots, D}I_d$.

\subsection{Bayesian inference}
Bayesian inference represents uncertainty over model parameters through probability distributions.  
Given data $\mathcal{D}$, the posterior distribution is obtained via Bayes’ rule~\citep{ML_Bishop_2006}:
\begin{equation}\label{bayes}
    p(\vect{v}|\mathcal{D}) = 
    \frac{p(\mathcal{D}|\vect{v})\, p(\vect{v})}{p(\mathcal{D})},
\end{equation}
where $p(\vect{v})$ is the prior, $p(\mathcal{D}|\vect{v})$ is the likelihood, and
$p(\mathcal{D})$ is the marginal likelihood obtained by integrating out $\vect{v}$.
Predictions for new inputs $\mat{X}_* \inCo{M \times D}$ are computed by marginalizing over the posterior:
\begin{equation}\label{bma}
    p(\vect{y}_*|\mat{X}_*, \mathcal{D}) 
    = \int p(\vect{y}_*|\mat{X}_*, \vect{v})\, p(\vect{v}|\mathcal{D})d\vect{v},
\end{equation}
where $\vect{y}_*$ is the vector of $M$ predicted outputs.

\section{Laplace Tensor Train Kernel Machine}\label{sec:LaplaceTTKM}
In this section, we introduce the Bayesian tensor train kernel machine and describe the components needed for its probabilistic formulation.

We start with the following discriminative model:
\begin{equation}\label{la_ttkm}
    \begin{cases}
        &y = f(\vect{x}, \vect{v}) + e = \vect{\phi}(\vect{x})^\top\vect{g}(\vect{v}) + e,\\
        &\vect{g}(\vect{v})_{\rBr{i_1\dots i_D}} = \mat{V}^{\rBr{1}}\rBr{i_1}\dots\mat{V}^{\rBr{D}}\rBr{i_D},\\
    \end{cases}
\end{equation}
where $\vect{g}(\cdot): \mathbb{C}^{I\sum_{i=1}^{D}R_{i-1}R_i} \rightarrow \mathbb{C}^{I^D}$ is the TT parameterization mapping, following the Definition~\ref{def_tt}, with $I \coloneq I_d$ for all $d=1\dots D$; 
$\vect{v} \coloneq \vecOp{\left[\vect{v}^{(1)}, \dots, \vect{v}^{(D)}\right]} \inCo{I\sum_{i=1}^{D}R_{i-1}R_i}$, with $\vect{v}^{(d)} \coloneq \vecOp{\tens{V}^{\rBr{d}}} \inCo{IR_{d-1}R_d}$; and 
$e \sim \mathcal{N}(0,\,\beta^{-1})$ is a Gaussian noise term with precision $\beta$. Based on the model formulation in Equation~\eqref{la_ttkm}, the conditional distribution of $y$ is given by:
\begin{equation}\label{cond_model_y}
p(y | \vect{x}, \vect{v}, \beta)
= \mathcal{N}\left(\vect{\phi}(\vect{x})^\top \vect{g}(\vect{v}), \beta^{-1}\right).
\end{equation}
Another key component of Bayesian inference is the prior distribution, which encodes initial beliefs about the parameters and influences the resulting model uncertainty. We specify the prior as
\begin{equation}\label{tt_weight_prior}
p(\vect{v}|\gamma) \coloneq \mathcal{N}(\vect{0}, \gamma^{-1}\mat{I}),
\end{equation}
where $\gamma$ is a precision hyperparameter.
To enable tractable Bayesian inference, we introduce a variational posterior and adopt a mean-field factorization:
\begin{equation}\label{q_mean_field}
q(\vect{v}, \beta, \gamma) = q(\vect{v})q(\beta)q(\gamma),
\end{equation}
which facilitates efficient approximation of the true posterior.

A central element of Bayesian inference is the posterior distribution over the model parameters~\eqref{bayes}, which reflects both model capacity and uncertainty. However, computing the posterior predictive distribution~\eqref{bma} requires evaluating a high-dimensional integral that is generally intractable. To address this, we employ the Laplace approximation~\citep{ML_Bishop_2006}, replacing the true posterior $p(\vect{v}|\mathcal{D}
)$ with a tractable Gaussian approximation $q(\vect{v})$, given by:
\begin{equation}\label{laplace}
    q(\vect{v}) = \mathcal{N}(\vect{v}|\hat{\vect{v}},\,\mat{C}_{\hat{\vect{v}}}),
\end{equation}
where the mean vector $\hat{\vect{v}}$ and the covariance matrix $\mat{C}_{\hat{\vect{v}}}$ are
estimated as follows:
\begin{equation}\label{loss}
    \begin{split}
    \mathcal{J}(\vect{v}) &\coloneq \dfrac{\beta}{2} \|\vect{y} -\mat{\Phi}\vect{g}(\vect{v})\|_2^2 + \dfrac{\gamma}{2}\|\vect{v}\|_2^2,\\
    \hat{\vect{v}} &\coloneq \arg\min_{\vect{v}}  \mathcal{J}(\vect{v}),\\
    \mat{C}_{\hat{\vect{v}}} &\coloneq \sqBr{\frac{\partial^2{ \mathcal{J}}}{\partial{\vect{v}}^\top \partial{\vect{v}}}}_{\vect{v} = \hat{\vect{v}}}^{-1}\text{ .}
     \end{split}
\end{equation}
The optimization problem in~\eqref{loss} is solved using the aforementioned ALS algorithm~\eqref{eq:als}. The main advantages of the algorithm, in the context of this paper, are discussed in the next section.

\subsection{One TT-core Bayesian model}
A critical and technically challenging aspect of the Laplace approximation is the computation of the covariance matrix $\mat{C}_{\hat{\vect{v}}}$, as defined in \eqref{loss}. In this work, we treat only a single TT-core as a Bayesian variable. This choice is supported both by theoretical considerations and by computational efficiency.
Within the Laplace framework, the approximation assumes that the mode of the posterior can be located accurately. This assumption does not hold when all TT-cores are treated as random variables and the optimization relies solely on gradient-based methods. In contrast, the ALS optimization procedure computes the optimal solution exactly with respect to one core at a time. When that same core is chosen as the Bayesian variable, the Laplace approximation is theoretically justified.

For these reasons, we restrict our analysis to the setting in which a single TT-core is modeled as a random variable, while the remaining cores are treated deterministically. This Bayesian treatment is aligned with prior work demonstrating that introducing uncertainty into only part of a model — such as the final layer of a neural network — can be both theoretically justified and empirically effective, see e.g. \cite{Last_Bayesian_Kristiadi_2020}.
Under this framework, the posterior distribution over the weights is given by:
\begin{equation}\label{eq_one_core}
    \begin{split}
        q(\vect{v}) &= \hat{q}(\vect{v}) \mathcal{N}(\vect{v}^{\rBr{d}}|\vect{\mu},\,\mat{C}), \\
        \mat{C} &= \beta\mat{A}^{\rBr{d}\top}\mat{A}^{\rBr{d}} + \gamma\mat{I},\\
        \vect{\mu} &= \rBr{\mat{A}^{(d)\top} \mat{A}^{(d)} + \dfrac{\gamma}{\beta} \mat{I}}^{-1}\mat{A}^{(d)\top}\vect{y},
    \end{split}
\end{equation}
where $\hat{q}(\vect{v}) = \rBr{\prod_{k=1, k\neq d}^{D}\delta(\vect{v}^{(d)} - \hat{\vect{v}}^{(d)})}$ and $\delta$ denotes the Dirac delta function.

\subsection{Predictive distribution}
Taking into account the chosen posterior approximation (applied to one TT-core), the predictive distribution is obtained by integrating the approximate posterior $q(\vect{v})$
over the conditional model~\eqref{cond_model_y}:
\begin{equation*}
    \begin{split}
    &p\rBr{\vect{y}_* | \mat{X}_*, \mathcal{D}} = \int_{\Omega_{\vect{v}}} p\rBr{\vect{y}_* | \mat{X}_*, \vect{v}} q\rBr{\vect{v}|\mathcal{D}} d\vect{v} \\
    & = \int_{\Omega_{\vect{v}}} p(\vect{y}_* | \mat{X}_*, \vect{v}^{\rBr{1}}, \dots, \vect{v}^{\rBr{D}}) \hat{q}(\vect{v}) \mathcal{N}(\vect{v}^{\rBr{d}}|\vect{\mu},\,\mat{C}) d\vect{v} \\
    & = \int_{-\infty}^{+\infty} \mathcal{N}(\vect{y}_*|\mat{A}^{\rBr{d}}_*\vect{v}^{\rBr{d}},\,\beta^{-1}\mat{I}) \mathcal{N}(\vect{v}^{\rBr{d}}|\vect{\mu},\,\mat{C}) d\vect{v}^{\rBr{d}} \\
    & = \mathcal{N}(\vect{y}_*|\mat{A}^{\rBr{d}}_*\vect{\mu},\,\beta^{-1}\mat{I} + \mat{A}^{\rBr{d}}_*\mat{C}\mat{A}^{\rBr{d}\top}_*),
    \end{split}
\end{equation*}
where $\mat{A}^{\rBr{d}}_*$ is the design matrix computed from the test inputs $\mat{X}_*$ and the optimized model parameters using~\eqref{eq:als}.

\textbf{Note:} If one assumes $\sigma_y^2 = \beta^{-1}$, $\mat{\Lambda} = \gamma^{-1}\mat{I}$, and $\mat{\Phi}\mat{W}_{\symbol{92} d} = \mat{A}^{\rBr{d}}$, then the resulting distribution coincides with the predictive distribution derived in~\cite{GPR_TT_2023}.

\subsection{Variational inference}\label{vi_hyperprior}
In a fully Bayesian treatment, it is standard to place priors over the hyperparameters — here, the precision terms $\beta$ and $\gamma$. Following~\cite{ML_Bishop_2006}, we model both using Gamma distributions:
\begin{equation}\label{prior_gamma_beta}
    p(z) = \operatorname{Gam}(a^{(0)}_z, b^{(0)}_z), \qquad z \in \{\beta, \gamma\}.
\end{equation}
The optimal variational factors are obtained via the coordinate ascent variational inference update rule:
\begin{equation}\label{alt_bishop}
    \ln q(\vect{z}_j) 
    = \mathbb{E}_{i \neq j}\!\left[\ln p(y, \vect{z})\right] + C,
    \qquad j = 1,2,3,
\end{equation}
where $\vect{z} \coloneq [\vect{z_1}, z_2, z_3]$, with $\vect{z}_1 \coloneq \vect{v}$, $z_2 \coloneq \beta$, $z_3 \coloneq \gamma$; $p(y, \vect{z}) = p(y|\vect{v}, \beta)p(\vect{v}|\gamma)p(\beta)p(\gamma)$ is the joint distribution; $q(\vect{z}_j)$ denotes the optimal variational distribution for the $j$-th variable, and the expectation is taken over all other variables. 

Applying Equation~\eqref{alt_bishop} to the noise precision $\beta$ results in the following variational Gamma posterior update for $t = 1, \dots, T$:
\begin{equation}\label{noise_precision}
    \begin{split}
    q(\beta) &= Gam(\beta|a_{\beta}^{(t)}, b_{\beta}^{(t)}), \\
    a_{\beta}^{(t)} &\coloneq a_{\beta}^{(t-1)} + \frac{N}{2}\text{, }\\
    b_{\beta}^{(t)} &\coloneq b_{\beta}^{(t-1)} + \frac{1}{2}\mathbb{E}_{\vect{v}}{\|\vect{y} -\mat{\Phi}\vect{g}(\vect{v})\|_2^2}.
    \end{split}
\end{equation}
Applying the same procedure yields the following variational update for the weight precision $\gamma$:
\begin{equation}\label{weights_precision}
    \begin{split}
    q(\gamma) &= Gam(\gamma|a_{\gamma}^{(t)}, b_{\gamma}^{(t)}), \\
    a_{\gamma}^{(t)} &\coloneq a_{\gamma}^{(t-1)} + \frac{DIR}{2}, \\
    b_{\gamma}^{(t)} &\coloneq b_{\gamma}^{(t-1)} + \frac{1}{2}\mathbb{E}{\vect{v}^\top\vect{v}}.
    \end{split}
\end{equation}
The posterior over the model weights $\vect{v}$ follows directly from Equations~\eqref{laplace} – \eqref{loss}, with the precision hyperparameters $\beta$ and $\gamma$ substituted by their expectations, $\mathbb{E}[\beta]$ and $\mathbb{E}[\gamma]$.

\section{Numerical Experiments}\label{sec:experiments}
To enhance training stability, we employ unit-norm polynomial features~\citep{SP_BTN_Mandic_2022} or Fourier features~\citep{VFF_Hensman_2017} as the local feature mapping $\vect{\phi}^{(d)}$ in~\eqref{tensor_product_features}.  
For simplicity in hyperparameter tuning, the number of basis functions is set uniformly across all modes, i.e., $I_d = I$ for $d = 1, \dots, D$. In addition, the TT-core treated in a Bayesian manner is chosen to match the core at which the ALS optimization is terminated, a choice motivated by the theoretical arguments discussed earlier.
The experiments in Sections~\ref{sec:ablation} and~\ref{sec:vi} use seven UCI regression datasets, chosen for their diversity in size and statistical properties, which is essential for a robust ablation study and for characterizing the behavior of the proposed LA-TTKM model.  
For each dataset, we split the data into 90\% (or 80\%) for training and 10\% (or 20\%) for testing.  
To assess and compare probabilistic methods — and different configurations thereof — we report the negative log-likelihood (NLL) and the root mean square error (RMSE) on the test set (lower is better). 
The NLL metric is computed as follows: 
\begin{equation*}
\text{NLL} = \frac{1}{M}\sum_{m=1}^M \left[ \dfrac{1}{2}\ln(2\pi v_m)
+ \frac{(y_m - \mu_m)^2}{2v_m} \right],
\end{equation*}
where $\mu_m$ and $v_m$ denote the predictive mean and variance for the $m$-th test point.
All experiments were conducted on a Dell Latitude 7440 laptop with a 13th Gen Intel Core i7-1365U CPU and 16 GB RAM, except for the experiment in Section~\ref{subsec:robot}, which required a full GP baseline. That experiment was run on a machine with 2 × AMD EPYC 7252 8-core CPUs and 256 GB RAM.
The full source code, including all implementation details and data required to reproduce the results, is available on GitHub.\footnote{https://github.com/AlbMLpy/laplace-ttkm}

\subsection{Ablation study}\label{sec:ablation}
In this set of experiments, we examine the key design choices underlying the LA-TTKM model, originally posed in~\cite{GPR_TT_2023}.  
First, we study which TT-core should be treated in a Bayesian manner and whether this choice depends on the TT-rank values $R_d$ for $d = 1, \dots, D-1$.  
Second, we investigate whether the ordering of input features affects model performance and, in particular, whether the Bayesian core should be associated with a specific feature $x_d$.
To address these questions, we evaluate the test NLL on seven real-world datasets while:  
(Q1) systematically varying the rank pattern, and  
(Q2) cyclically shifting the input features.  
All remaining hyperparameters are fixed so that the total number of model parameters remains smaller than the number of training samples.

\subsubsection{\textbf{Q1: Which TT-core should be Bayesian?}}
\textit{The results in Table~\ref{table:bayes_tt_core} indicate that any TT-core can be chosen as the Bayesian core, and this choice does not depend on the TT-rank pattern.}  
In the table, each row corresponds to a dataset and each column to a different rank configuration.  
\textit{Pattern~1} denotes TT-ranks of the form $[R, R, R]$ with $R > 1$;  
\textit{Pattern~2} uses $[R, P, R]$ with $P > R$;  
and \textit{Pattern~3} uses $[R, P, R]$ with $P < R$.  
These patterns illustrate the general idea; in practice, the number of TT-ranks depends on the dimensionality of each dataset.
The table entries are interpreted as follows:  
for example, in the Boston dataset under \textit{Pattern~1}, the entry "5/13" indicates that the dataset has 13 dimensions (i.e., 13 TT-cores), and that the 5th core yields the best test NLL among all 13 possible choices of Bayesian core.  
Table~\ref{table:bayes_tt_core} shows that changing the TT-rank pattern has minimal effect on which core performs best across datasets. This further indicates that the choice of Bayesian core is largely independent of the rank configuration.

\begin{table}[t]
\centering
\caption{“Which TT-core should be Bayesian?” An entry of the form $a/c$ means that TT-core $a$ yields the best result out of the $c$ cores.}
\label{table:bayes_tt_core}
\begin{tabular}{||l||c|c|c||}
\toprule
Data & Pattern 1 & Pattern 2 & Pattern 3 \\
\midrule
Boston & 5/13 & 7/13 & 9/13 \\
Concrete & 5/8 & 5/8 & 2/8 \\
Energy & 5/8 & 5/8 & 5/8 \\
Kin8nm & 8/8 & 8/8 & 8/8 \\
Naval & 9/16 & 12/16 & 9/16 \\
Protein & 9/9 & 9/9 & 9/9 \\
Yacht & 1/6 & 1/6 & 1/6 \\
\bottomrule
\end{tabular}
\end{table}

\subsubsection{\textbf{Q2: Which feature $x_d$ should be Bayesian?}}

\textit{Table~\ref{table:bayes_tt_features} suggests that there is no universally optimal choice.}  
Each column in the table corresponds to a cyclic shift of the input features.  
For example, given an input vector $\vect{x} = [x_1, x_2, \dots, x_D]$,  
the \textit{No Shift} configuration uses the features as given.  
\textit{Shift 1} applies the permutation  
$\vect{x} = [x_D, x_1, \dots, x_{D-1}]$, and higher shifts follow the same logic.  
The table entries are interpreted in the same way as in the previous experiment.
Across most datasets, the best-performing Bayesian core tends to be either the first or the last TT-core.  
This behavior is expected: these cores contain the fewest parameters since $R_0 = R_D = 1$,  
and Bayesian averaging over smaller parameter spaces typically yields more stable predictions.

To summarize, these experiments lead to two key conclusions.  
First, the TT-rank pattern has negligible effect on Bayesian core performance,  
which allows the use of a single uniform rank and substantially simplifies hyperparameter tuning.  
Second, the choice of feature ordering does not practically influence performance,  
meaning that additional feature permutations are unnecessary.  
For computational efficiency and improved scalability, we therefore use the \textit{first} TT-core as the Bayesian core, as it contains the smallest number of parameters.

\begin{table}[t]
\centering
\caption{“Which feature $x_d$ should be Bayesian?” An entry of the form $a/c$ means that TT-core $a$ yields the best result out of the $c$ cores.}
\label{table:bayes_tt_features}
\begin{tabular}{||l||c|c|c|c||}
\toprule
Data & No Shift & Shift 1 & Shift 2 & Shift 3 \\
\midrule
Boston & 4/13 & 1/13 & 2/13 & 3/13 \\
Concrete & 5/8 & 5/8 & 3/8 & 2/8 \\
Energy & 8/8 & 8/8 & 1/8 & 1/8 \\
Kin8nm & 8/8 & 8/8 & 8/8 & 8/8 \\
Naval & 16/16 & 16/16 & 16/16 & 16/16 \\
Protein & 9/9 & 9/9 & 9/9 & 9/9 \\
Yacht & 1/6 & 6/6 & 6/6 & 6/6 \\
\bottomrule
\end{tabular}
\end{table}

\subsection{Variational inference for hyperparameters}\label{sec:vi}
In this section, we compare the LA-TTKM model equipped with variational inference (VI) to a version that selects the hyperparameters $\gamma$, $\beta$ from the set $\{0.001, 0.01, 0.1, 1, 10\}$ via cross-validation (CV) in a regression setting.
To ensure a fair comparison, parallelization is disabled for both methods.
Each algorithm is trained on several datasets, and the resulting test NLL and training times are summarized in Table~\ref{table:vi_cv}.
The reported means and standard deviations are computed over 10 independent restarts.

The results indicate that VI consistently achieves faster training times than CV, providing an average speed-up of approximately 65\%.
In terms of predictive performance, the outcomes are dataset-dependent; however, VI performs comparably to CV in most cases, as reflected by similar mean NLL values and overlapping standard deviation intervals.

\begin{table}
\centering
\caption{
Test NLL results and training time of the LA-TTKM model using VI or CV.}
\label{table:vi_cv}
\scalebox{0.95}{\begin{tabular}{||l||c|c||c|c||}
\toprule
 & \multicolumn{2}{|c||}{NLL Test $\downarrow$} & \multicolumn{2}{|c||}{Time (s) $\downarrow$} \\
 \midrule
Data & CV & VI & CV & VI  \\
\midrule
Boston & 1.16±0.05 & 1.55±0.83 & 159.6±7.4 & 1.8±0.4 \\
Concrete & 0.38±0.09 & 0.26±0.14 & 1397.5±63.0 & 48.6±11.5 \\
Energy & -0.07±0.01 & 0.93±0.62 & 119.8±8.5 & 1.4±0.5 \\
Kin8nm & 0.51±0.19 & 0.38±0.02 & 146.4±8.8 & 2.7±0.5 \\
Naval & -0.13±0.02 & -1.16±0.15 & 237.0±1.8 & 6.7±0.5 \\
Protein & 1.23±0.00 & 1.13±0.01 & 354.9±19.1 & 15.5±0.6 \\
Yacht & 0.03±0.22 & -0.41±0.09 & 126.3±13.5 & 0.9±0.4 \\
\bottomrule
\end{tabular}}
\end{table}

\subsection{Inverse dynamics of a robotic arm}\label{subsec:robot}
\begin{figure}
\begin{center}
\includegraphics[width=8.4cm]{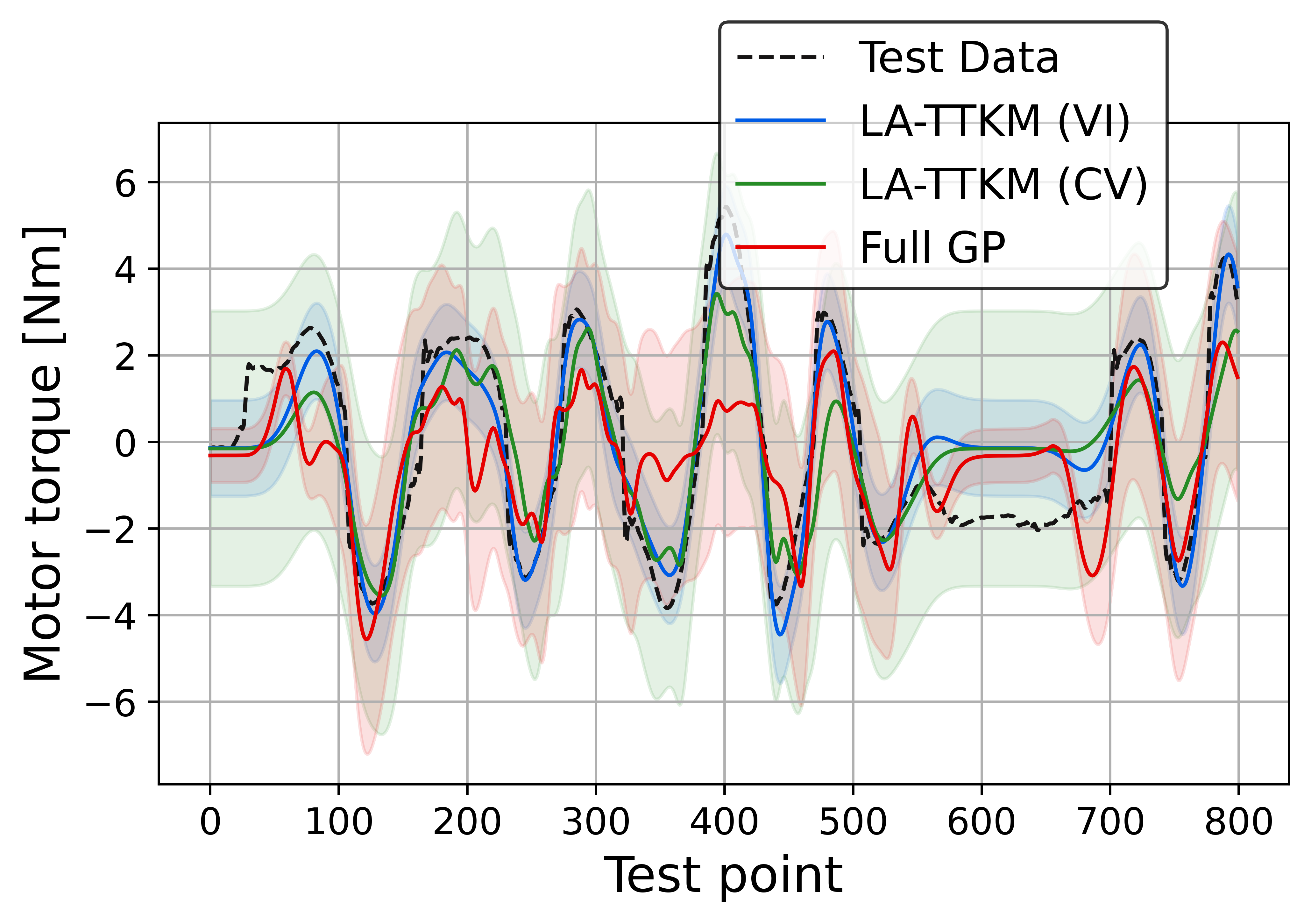}    % The printed column width is 8.4 cm.
\caption{Predicted torques for a single joint. Solid lines denote the predictive mean, and shaded regions represent the $\pm$1 standard deviation uncertainty band.} 
\label{fig:robot_dynamics}
\end{center}
\end{figure}

In this experiment, we evaluate the LA-TTKM model with variational inference on a nonlinear system identification task. We use the large-scale industrial robot dataset from~\cite{Robot_Data_2022}, containing
$N = 39~988$ training samples and $M = 3636$ test samples.
The objective is to learn the inverse dynamics of a robotic arm. Each input is an 18-dimensional vector comprising joint positions, velocities, and accelerations, while the output is a 6-dimensional vector of joint torques. In this study, we predict a single torque component.
For LA-TTKM, we set $R_1=\dots=R_{D-1}=3$ and use $I=8$ basis functions per dimension. Without the low-rank tensor structure, the model would contain $8^{18}$ parameters. We compare VI and CV against a full GP baseline, with results summarized in Table~\ref{table:robot_dynamics} and Figure~\ref{fig:robot_dynamics}.
The results show that VI provides substantial computational benefits, achieving a 60× speedup over CV and a 325× speedup over full GP. In terms of predictive accuracy, VI yields lower RMSE than both baselines, while exhibiting higher NLL due to occasionally overconfident uncertainty estimates, as seen in the narrow predictive intervals in Figure~\ref{fig:robot_dynamics}.

\begin{table}
\centering
\caption{Trade-off between predictive performance (NLL, RMSE on test) and training time for LA-TTKM using VI, CV and, Full GP}
\label{table:robot_dynamics}
\begin{tabular}{||l|c|c|c||}
\toprule
Model & NLL $\downarrow$ & RMSE $\downarrow$ & Time (s) $\downarrow$ \\
\midrule
LA-TTKM (VI) & 6.2 & 1.1 & 22.8 \\
LA-TTKM (CV) & 2.6 & 1.4 & 1364.2 \\
Full GP & 4.4 & 1.8 & 7446.3 \\
\bottomrule
\end{tabular}
\end{table}

\section{Conclusion}
In this work, we developed a Bayesian tensor train kernel machine (LA-TTKM) using a Laplace approximation applied to a selected TT-core, combined with variational inference.
Our principled probabilistic framework allowed us to resolve several open questions posed in~\cite{GPR_TT_2023}, including which TT-core should be treated as Bayesian, the influence of TT-ranks on Bayesian modeling, and efficient hyperparameter selection.
Experimental results showed that the choice of Bayesian core is largely independent of both features and ranks, suggesting that, for scalability, the first TT-core can be chosen for this role.
Furthermore, the proposed variational inference scheme achieves predictive performance comparable to cross-validation while reducing training time by approximately 65× on average.
For future work, we plan to explore richer probabilistic formulations, such as making multiple—or all—TT-cores Bayesian or employing mean-field approximation.
Another promising direction is the automatic adaptation of TT-ranks, allowing the model to grow incrementally as needed without retraining from scratch.

\begin{ack}
This publication is part of the project Sustainable learning for Artificial Intelligence from noisy large-scale data (with project number VI.Vidi.213.017) which is financed by the Dutch Research Council (NWO).
\end{ack}

%\section*{DECLARATION OF GENERATIVE AI AND AI-ASSISTED TECHNOLOGIES IN THE WRITING PROCESS}
%During the preparation of this work the author(s) used [NAME TOOL / SERVICE] in order to [REASON]. After using this tool/service, the author(s) reviewed and edited the content as needed and take(s) full responsibility for the content of the publication.

\bibliography{ifacconf}

%\appendix
%\section{A summary of Latin grammar}    % Each appendix must have a short title.
%\section{Some Latin vocabulary}              % Sections and subsections are supported  
\end{document}